\documentclass[aps,twocolumn,floats,pre]{revtex4-1}
\usepackage{graphics,graphicx,epsfig,color}
\usepackage{amssymb,amsmath}
\usepackage{epsf,wrapfig}
\usepackage[mathcal]{euscript}

\usepackage{units} 
\usepackage{commath} 
\usepackage[]{titlesec}
\titleformat{\section}[block]{\color{black}\bfseries\filcenter}{}{1em}{}
\usepackage{hyperref}
\usepackage{upgreek} 

\usepackage{multirow}

\usepackage{subcaption}
\usepackage[font=small,labelfont=bf,justification=raggedright, format=plain]{caption} 
\titlespacing\section{0pt}{10pt plus 2pt minus 0pt}{10pt plus 2pt minus 2pt}

\newcommand{\figtitle}[1]{\textbf{#1}}

\begin{document}

\title{Pixel personality for dense object tracking in a 2D honeybee hive}


\title{Pixel personality for dense object tracking in a 2D honeybee hive}

\author{Katarzyna Bozek\textsuperscript{\,a},
Laetitia Hebert\textsuperscript{\,a},
Alexander S Mikheyev\textsuperscript{\,a} \&
Greg J Stephens\textsuperscript{\,a,b} 
} 

\affiliation{
\textsuperscript{a}Okinawa Institute of Science and Technology, 1919-1 Tancha Onna-son, Kunigami-gun, Okinawa 904-0495, Japan\\
 \textsuperscript{b}Department of Physics and Astronomy, VU University Amsterdam, 1081 HV Amsterdam, The Netherlands
}


\begin{abstract}
Tracking large numbers of densely-arranged, interacting objects is challenging due to occlusions and the resulting complexity of possible trajectory combinations, as well as the sparsity of relevant, labeled datasets. Here we describe a novel technique of collective tracking in the model environment of a 2D honeybee hive in which sample colonies consist of $N\sim10^3$ highly similar individuals, tightly packed, and in rapid, irregular motion. Such a system offers universal challenges for multi-object tracking, while being conveniently accessible for image recording. We first apply an accurate, segmentation-based object detection method to build initial short trajectory segments by matching object configurations based on class, position and orientation. We then join these tracks into full single object trajectories by creating an object recognition model which is adaptively trained to recognize honeybee individuals through their visual appearance across multiple frames, an attribute we denote as pixel personality. Overall, we reconstruct $\sim46\%$ of the trajectories in $5 \, {\rm min}$ recordings from two different hives and over $71\%$ of the tracks for at least $2 \, {\rm min}$. We provide validated trajectories spanning $3,000$ video frames of $876$ unmarked moving bees in two distinct colonies in different locations and filmed with different pixel resolutions, which we expect to be useful in the further development of general-purpose tracking solutions.
\end{abstract}
                                                                                                                                                                                                                                                                                                                                  
\keywords{}
\maketitle

\section*{Introduction} 
From cells in tissues to human crowds, the tracking of densely packed, active objects is a challenging and important problem.  Indeed, the understanding of collective phenomena in biology across scales -- from cellular interactions to the organization of animal groups -- requires efficient tracking methods to connect ensemble behavior with the mechanisms of individual components (see e.g.~\cite{Crall2018-dk,Mersch2013-uf}).  The automated tracking of individual objects in dense groups based on video recording is also of practical interest for the implementation of monitoring frameworks.

Social insects are an important category of animal groups whose reproduction and survival depend on the intrinsic organization and cooperation of the entire colony.  Honeybees, an example of social insects, live in large organized family groups and perform activities unavailable to solitary insects, such as nest construction, division of labor, or group defense \cite{Seeley2009-yz}. Due to long-established human husbandry, bees are accessible for observation with the use of an \emph{observation beehive}. In such a hive the entire colony is placed on one surface of a honeycomb covered with glass. Tight spacing between the glass and the comb constrains the colony to an approximately 2D environment in which colony activities are performed by thousands of individuals in dense configurations and constant motion. The complexity of tracking within the observation hive environment is representative of other systems and thus of general interest.

We propose a new method for tracking of a large number of highly similar individuals in dense configurations. We adapt a previous method for identifying bee position and orientation \cite{Bozek2017-sj} by retraining with samples from our imaging arrangement.  We next create initial track fragments with a basic matching approach using the difference in object class, orientation and position across neighboring frames. Finally, we iteratively learn visual features of hundreds of individuals within the short tracks, effectively assigning a unique ``pixel personality'' to each object. We use these learned visual features to match track fragments belonging to the same individual into full trajectories spanning the entire recording. 

With the remarkable rise of deep learning, there has been surprisingly little progress with multi-target tracking, in part due to the lack of sufficiently large and annotated datasets. As an outcome of our work,  we also provide a unique dataset of a large number of trajectories in a dense and challenging environment with nearly identical objects.
\section*{Related work}
Among the diverse applications of multi-object tracking, crowd tracking has received the broadest attention in computer vision due to applications in surveillance systems, robotics, and human-computer interaction environments. Given the difficulty of tracking multiple moving, interacting, and visually overlapping targets, numerous approaches have explored the spatiotemporal patterns within human crowds and used local motion correlations to aid the tracking task \cite{Kratz2010-kb,Ali2009-sf,Ge2012-tz,Rodriguez2011-qo}.

Expanding tracking beyond locally correlated groups necessitates the use of additional cues to prevent identity swaps. Examples include temporal and spatial linking through quadratic programming \cite{Henschel2016-zn}, using appearance and motion cues \cite{Bing_Wang2017-xf}, blending appearance to multiple hypothesis tracking \cite{Kim2015-hr}, or integrating motion, time, position, size, and appearance into a single affinity model \cite{Kuo2011-ie}. With the broad adoption of deep learning for image analysis, convolutional neural networks (CNNs) have opened up new possibilities of extracting visual cues in the task of multi-object tracking. In particular, siamese architectures have attracted broad interest for their ability of associating detection pairs \cite{Leal-Taixe2016-wn}.

Despite the capacity of recurrent neural networks (RNNs) to learn temporal patterns, few studies have shown their application to multi-object tracking. However, a notable example \cite{Milan2016-gq} suggested an integrated architecture for trajectory association including object birth and death based on existing detections. Trained end-to-end, this model has shown high performance in simulated data and crowd recordings. RNNs have also been applied to space occupancy prediction based on unsupervised representation learning in the context of multi-object tracking \cite{Ondruska2016-xy,Ondruska2016-dd}. 

The need to create standards for the evaluation of tracking methods has inspired the creation of the Multi-Object Tracking benchmark, with both human tracks and also elements of scene and other object classes introduced in the benchmark's 2016 release (MOT16) \cite{MOT16}. This dataset includes a total of $1,276$ tracks within $14$ recordings of up to $1,000$ frames, addressing major challenges in scene tracking such as varying conditions, viewpoints, and a moving camera. 

Compared to the MOT16 benchmark, honeybee colonies contain a significantly higher number of individuals ($\sim1,000$ each), a number prohibitively large for the comprehensive exploration of track association hypotheses. Individual bees are generally of the same size, largely similar, and not available for recording in isolation, which limits the usefulness of comprehensive appearance models \cite{Romero-Ferrero2018-fb}. The beehives also exhibit a varied, uneven background composed of comb cells, brood, and food, the viewpoint of the recordings is static, and object classes are limited to two types of bee body postures.

Due to the difficulties of multi-object tracking, studies in biological systems often produce collective recordings under controlled laboratory conditions or with the marking of individuals. Typical experimental conditions involve plain 2D or 3D arenas with uniform illumination and high foreground-background contrast \cite{Kabra2013-mg,Romero-Ferrero2018-fb}. The controlled aspect of these arrangements allows for accurate object detection as well as extraction of comprehensive appearance cues from large volumes of recordings of each individual \cite{Romero-Ferrero2018-fb}. The marking of individuals with unique barcodes has also been extensively used in the study of social insects \cite{Mersch2013-uf,Wario2015-fm,Gernat2018-kw}, but has important feasibility limitations in large colonies with short generation time spans. Both the laboratory conditions and marking itself can also have an important effect on behavior, necessitating versatile methods for marker-less tracking of groups of individuals in their natural environments. Such methods would also create the opportunity for collective tracking in systems that cannot be observed under controlled conditions and that cannot be easily marked.

\section*{Data}

We collected two video recordings from observation beehives in two different locations and filmed with cameras of resolution ($5120\times5120\,{\rm pixels}$ and $3860\times2160\,{\rm pixels}$, respectively) at $30\,{\rm fps}$ (Fig.~S1) for $5\,{\rm min}$. The images of the first recording were downsized by a factor of two along both axes to reach a similar spatial resolution of the second recording, in which a bee's largest dimension is $\sim 80\,{\rm pixels}$. 

\section*{Object detection}

We followed and improved a previous segmentation approach for the efficient estimation of honeybee position and orientation \cite{Bozek2017-sj} by combining the orientation and position loss functions into one architecture and adding one convolutional layer (Fig.~S2). This modified network network enables the simultaneous computation of both object orientation and class. 

We used two object classes corresponding to a fully visible bee and an abdomen of a bee which is inside of a comb cell. For the abdomen class, the orientation angle is not defined and is set to $0$. We retrained the modified architecture on a previous dataset \cite{Bozek2017-sj}, downsized by a factor of two on both axes to match the resolution of our video recordings.  We used ellipse-shaped segmentation labels for marking fully visible bees and round-shaped for marking abdomens of bees inside comb cells. Both labels were centered over reported bee central point position, ellipse-shaped labels were additionally rotated to align with the bee body axis. We resized the segmentation labels to $r1 = 6.7\,{\rm pixels}$ and $r2 = 11.7\,{\rm pixels}$ for the semi-minor and semi-major axes of the ellipse-shaped segmentation label of a fully visible bee and $r = 6.7\,{\rm pixels}$ radius for the round-shaped segmentation label of bee abdomens inside the comb cells (Fig.~S3). This resizing produced more precise position estimation and improved bee separation in dense configurations. The segmentation output was post-processed as described previously \cite{Bozek2017-sj}.

To improve the accuracy of the detection model, which was trained on the original data \cite{Bozek2017-sj}, we retrained the network on a small number of frames from our video recordings.  We performed the labelling n an iterative manner, where predictions of the detection model were first manually corrected using a custom interface (Fig.~S4), and then used to retrain the detection model. In each iteration, two frames of each recording spaced by $0.1\,{\rm s}$ were corrected and the network was trained for $5$ iterations on image patches of $256\times256\,{\rm pixels}$ covering the newly labeled frames. In each iteration we noticed a decrease by two of the time spent on labeling corrections of the following two frames and we stopped the process when no time gain remained, which happened when $32$ frames of each recording were labeled. The final network used for object detection was retrained for $30$ iterations on the newly labeled set of frames. 

\begin{figure}[t]
  \centering
    \includegraphics[width=0.46\textwidth]{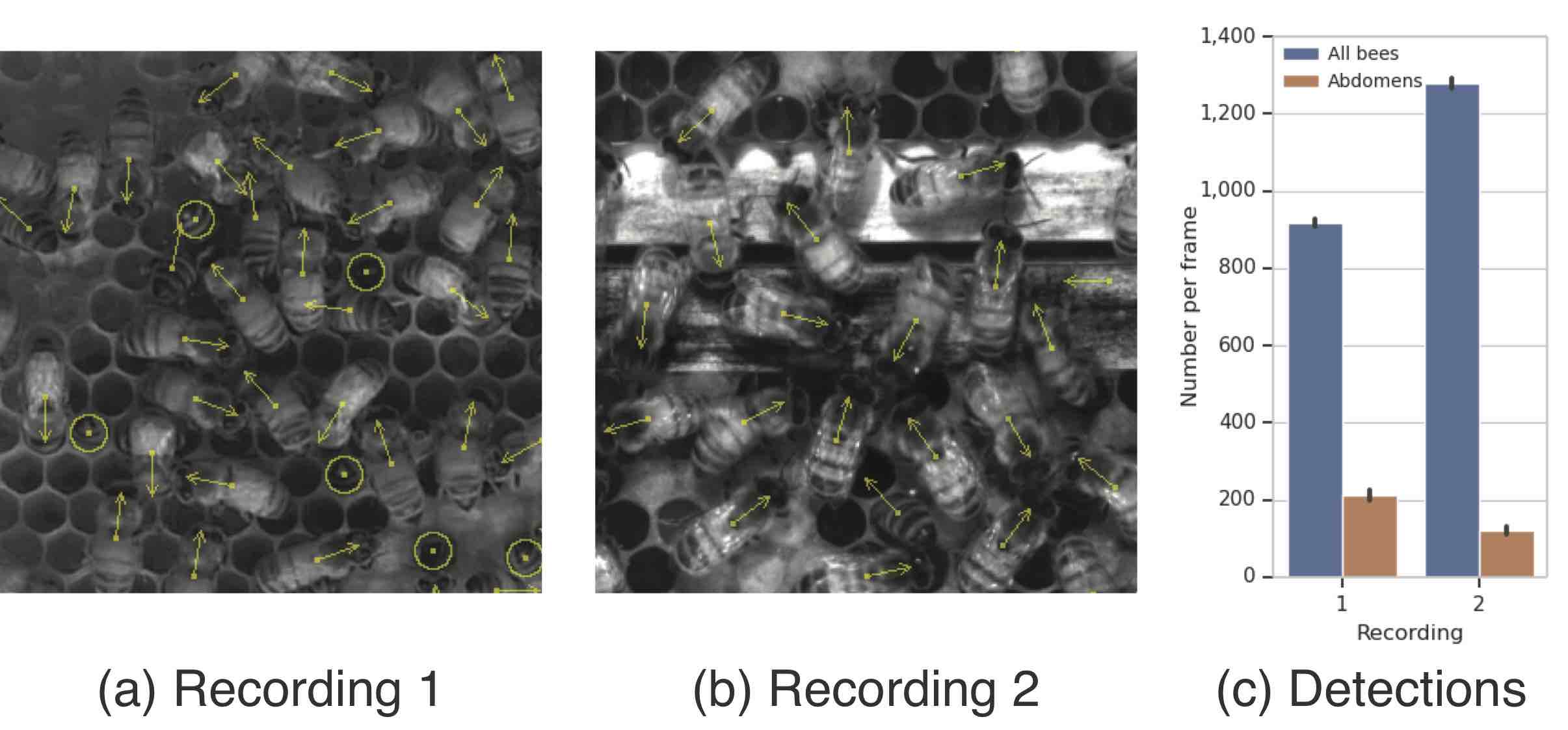}
    \caption{{\bf Reliable object detection is the foundation of our dense tracking approach.} (a,\,b) Example detections from two recordings. Body centers and head-tail orientation are marked with arrows, abdomens of bees inside comb cells are marked with circles. (c) Average number of detections per frame with vertical black lines denoting the standard deviation. The low variation in the number of detected objects is consistent with a constant and small false positive rate of $\sim0.06$ \cite{Bozek2017-sj}.} \label{fig:1detect} 
\end{figure}

The training of the segmentation network and the detection of bees were performed on image patches of size $256\times256\,{\rm  pixels}$. Given the higher error rates of object detection along the image boundaries \cite{Bozek2017-sj}, we overlapped the patches by a margin of $25\,{\rm  pixels}$ and results within margins were discarded. Both recordings captured images at $30\, {\rm fps} $ and we used a segment of $5\,{\rm  min}$ downsampled to $10\,{\rm fps}$. Our segmentation procedure resulted in $2,756,532$ and $3,840,312$ detections from recording 1 and 2, respectively with accuracy (after removing image margins) of $\sim0.06$ false positive rate, an average position error of $4.9\, {\rm pixels}$ and an orientation angle error of $9.7^\circ$ (Fig.~\ref{fig:1detect}), thus providing a strong foundation for trajectory reconstruction.

\section*{Constructing track fragments using configuration-based matching}

For each frame, the detection algorithm provides a list of entries $x, y, c, \alpha$, where $x, y$ are object coordinates, $c$ is the object category (fully visible bee or an abdomen only), and $\alpha$  is the orientation angle. We combine these four features into a similarity measure $D_{ij} $ to identify object detections across neighboring frames,
\begin{equation}
\label{eq-distance}
D_{ij}=d_{ij}+w|c_i - c_j| + w|\sin(\alpha_i - \alpha_j)| + |d_{x_{ij}}| + |d_{y_{ij}}|,   \nonumber \\
\end{equation}
where $d_{ij}$ is the Euclidian distance between objects $i$ and $j$, $w$ is a scaling factor of the class and angle differences between the objects, and $d_{x_{ij}}$ and $d_{y_{ij}}$ are differences in the motion vectors along the $x$ and $y$ axis in this matching step compared to the previous one of the same objects $i$ and $j$. Since the orientation angle is based on the body axis estimation \cite{Bozek2017-sj}, the angle component of $D_{ij} $ penalizes deviations from axis orientation rather than the head-tail orientation. We set the scaling factor $w=20$. 

\begin{figure}[h]
  \centering
    \includegraphics[width=0.5\textwidth]{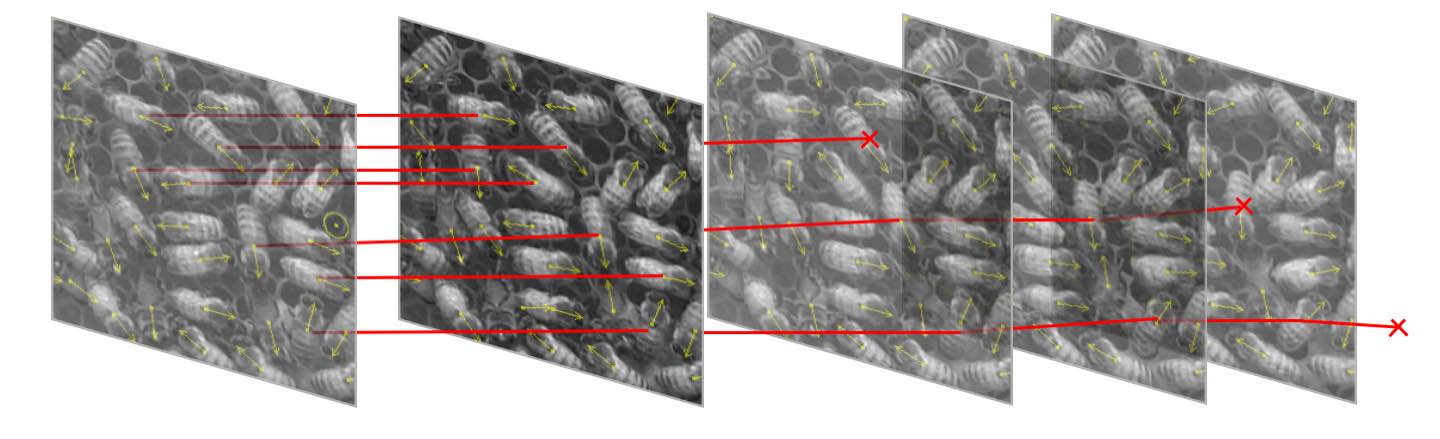}
    \caption{{\bf Object detections are joined into short track fragments using a simple distance metric.} We join the nearest detections in pairs of consecutive video frames using a distance measure composed of position, orientation angle, object class and direction of movement. The joining procedure results in short track fragments of varying lengths as well as start and end points.}\label{fig:2matching}
\end{figure}

Using $D_{ij} $, objects in the following frames of each recording were matched in an iterative manner, Fig.~\ref{fig:2matching}. Starting from lowest values of $D$ between two consecutive video frames the respective objects were matched into a track, until no pairs of objects with $D < 50$ remained. Matching was done between following pairs of frames and tracks for which no match was found over the last $10$ frames were ended. Only tracks of length $> 30$ were kept as shorter trajectories were considered as individual detections. This greedy detection matching is reminiscent of the Hungarian algorithm, however we do not search for a complete matching as occlusions and detection errors can lead to wrong associations. We only join detections that are sufficiently close which ensures that the track fragments are correct and thus leaving the task of expanding these fragments to the appearance-based, pixel personality matching described below.

The matching procedure resulted in $8,895$ and $8,565$ track fragments in recordings 1 and 2, respectively. The fragments have various lengths (Fig.~S5) and span a total of $\sim 10^6$ detections in each recording. Within the first $30\,{\rm s}$ of the recording we searched for a frame with the highest number of tracks with length $> 100$. These tracks represented the initial track set that was extended further as described in the next section. This set contains $794$ initial trajectories from recording 1 and $1,115$ in recording 2.  We also note that $36$ and $137$ trajectories respectively, spanning $> 95\%$ of frames, were considered complete and not extended further.

\begin{figure*}[t]
  \centering
    \includegraphics[width=.8\textwidth]{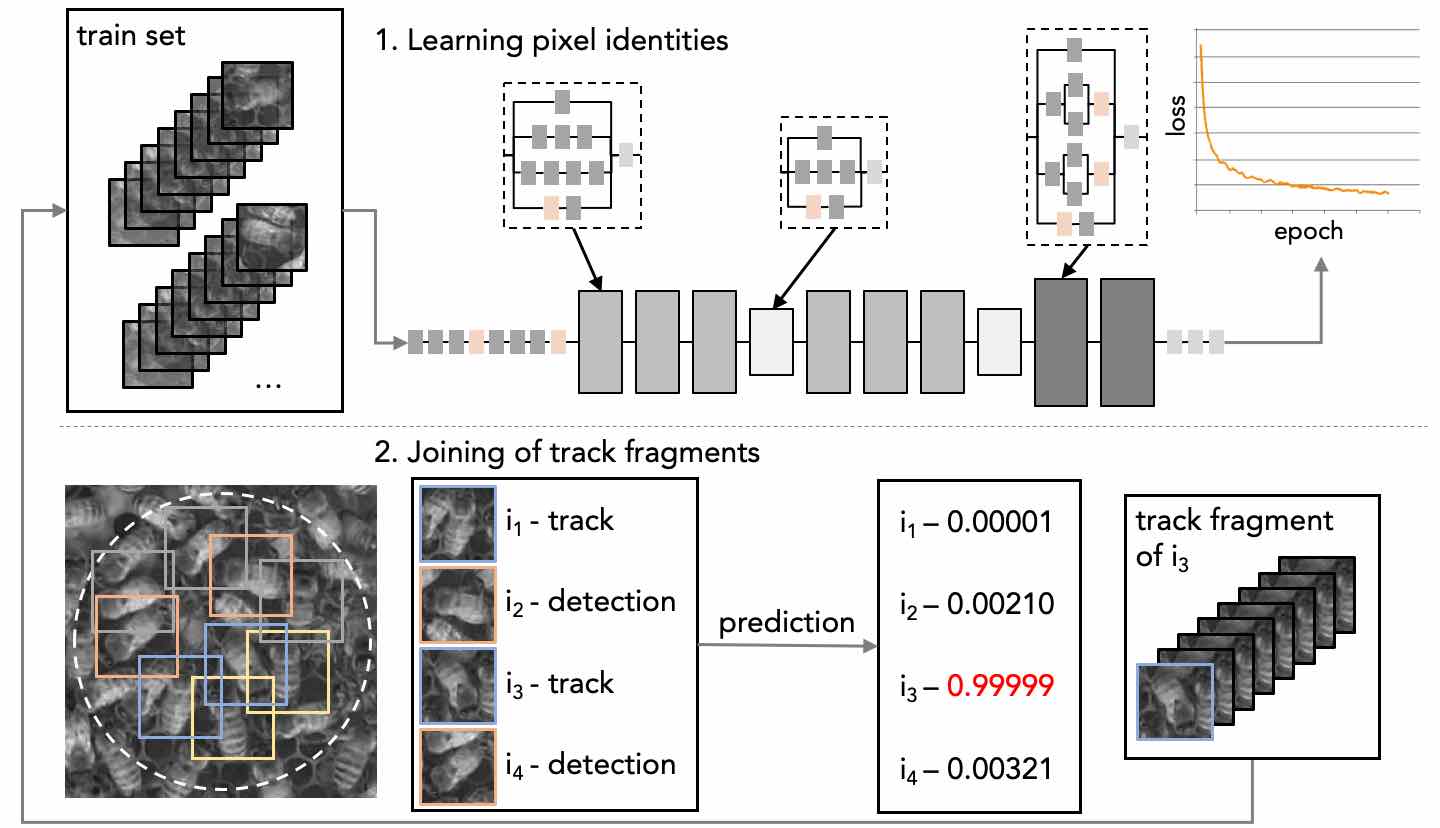}
    \caption{\figtitle{Overview of the track joining algorithm.} (upper left) Initial track fragments from different individuals are used as a training set for the object identity recognition network \cite{Szegedy2015-xm}. After the training loss function decreases below a predefined value, we consider track extensions through the joining of fragments. For a given object we examine the video frame immediately following the current end of the assembled trajectory and consider all detections within a predefined similarity (bottom-left, dashed line) from the last point in the trajectory. Among these detections, some belong to  trajectories of other objects (yellow squares), some are part of a track and are not at the beginning of track (gray squares), some are part of a track and are at the beginning of this track (blue squares), and some represent individual detections (single-frame fragments, orange squares). For detections belonging to the latter two categories we use the Inception V3 network to predict detection identity. If the predicted identity of any of the tested detections matches the given object (here, $i_3$) this detection is added to the trajectory. If the chosen detection is at the beginning of a track fragment, all detections belonging to this fragment are added to the assembled trajectory. Finally, the network is retrained on the updated training set and the process repeats.}\label{fig:3recognition}
\end{figure*}

\section*{Joining track fragments using appearance-based matching}

The track segments obtained using the matching criteria defined above are fragmented due to imperfect detections, occlusions, and irregular, sometimes rapid motion. To connect the fragmented tracks and individual detections into trajectories spanning the entire recording, we devised a matching approach exploiting appearance, space, and temporal cues, as well as taking into account the space occupancy of other tracked objects.

We captured cues to individual identity across varying visual appearances, an attribute we denote as pixel personality, by using the object recognition network Inception V3 \cite{Szegedy2015-xm}. We also note previous approaches that exploit subtle but consistent visual differences for multi-animal tracking \cite{Perez-Escudero2014-se,Romero-Ferrero2018-fb}. The initial track segments described above were used as a training set for learning object identities. Images of individuals from each track were cropped to the size $80\times80\, {\rm pixels}$ centered over the detected positions of each individual. Only $N$ occurrences from the latter part of the individual's track were incorporated into the training set. If a trajectory was shorter than $N$, the occurrences were repeated until size $N$ was obtained. These individual instances were stored in memory, preserving their order in the recording. We also randomly sampled images of background from the space in between the bee detections from different video frames. We collected $10^4$ background images and selected $N=250$ which allows for feasible processing time. 
\begin{table*}
\begin{center}
\begin{tabular}{|l|c|c|c|c|c|c|c|c|c|c|}
\hline
Recording & MT2$\uparrow$ & MT5$\uparrow$ &  ML2$\downarrow$ & ML5$\downarrow$ & Det. err. & ID swaps & Occlusion & Lost & Assoc. MT2 & Assoc. MT5 \\
\hline\hline
Rec. 1 & 0.74 & 0.46 & 0.01 & 0.12 & 0.03 & 0.19 & 0.10 & 0.17 & 981 & 10,681 \\
Rec. 2 & 0.71 & 0.46 & 0.02 & 0.16 & 0.02 & 0.30 & 0.03 & 0.10 & 470 & 3,772 \\
\hline
\end{tabular}
\end{center}
\caption{\figtitle{Summary results.} The tracking quality metrics shown are: trajectories tracked for at least $1\ min\ 40\ sec$ (MT2), for at least $4\ min$ (MT5), for less than $10\ sec$ (ML2), less than $30\ sec$ (ML5). Within the trajectories not counted as MT5 sources of errors were categorized as detection error (\emph{Det. err.}), identity swap (\emph{ID swaps}), occlusion (\emph{Occlusion}), or trajectory loss due to other reasons, for example, lack of detections or lack of detections recognized as the respective object (\emph{Lost}). Columns \emph{Assoc. MT2} and \emph{Assoc. MT5} list the number of track and detection associations performed using the object recognition network described previously.}
\label{tab:results}
\end{table*}

Starting with our assembled training set, including background images and images from separate objects labeled as a distinct category, we iteratively performed training and matching steps, Fig.~\ref{fig:3recognition}. In the train step, the network is trained until the loss decreases below a selected value ($L<0.01$). In addition, two data augmentation operations are randomly performed on the train set -- masking and flipping. Masking is aimed at reducing the background influence on the object recognition and consists of covering the image's outer parts and leaving only the center either square- or round-shaped (Fig.~S6). Flipping is performed along the x- and y-axes. The loss function of background images is down-scaled by a factor of $0.1$ to compensate for their higher number compared to other object categories. 

After the network reaches a predefined loss $L$, the matching step is performed. The object identities learned by the network are used to find the most probable object detections in the video frame following the frames of the given object instances contained in the train set. Specifically, for a given trajectory $k$ that ends at frame $t$ at position $p$, we search for detections in the frame at $t+d_t$ which fulfill the following criteria: (1) are not already assigned to another object, (2) are not part of a track fragment or are within first 10 instances of a track fragment, (3) are no further than a distance $D \sqrt{d_t}$ from $p$, where $D=80\,{\rm pixels}$ is approximately the largest object dimension size and $d_t$ is the difference (in frames) between currently tested frame and the last frame of trajectory $k$ (Fig.~\ref{fig:3recognition}). Detections fulfilling all three criteria are cropped from the video frame and tested with the object recognition network for their similarity to the object in trajectory $k$. 

From the set of tested detections, the one showing the highest softmax score that is above a cutoff score of $s=0.1$ in the object category $k$ is selected for extending trajectory $k$. This detection is added to the trajectory and the train set with label $k$. The earliest instances of trajectory $k$ are removed from the train set keeping the number of instances of label $k$ at the same number $N$.  If this detection belongs to a track fragment (recall that the detection can be at most 10 frames from the start of a track fragment), the entire track is added to the train set with the label $k$ and the same number of earlier instances of this trajectory are removed from the train set. If no tested detection shows a score above $s$ or no detection is found fulfilling the location and occupancy criteria, trajectory $k$ is not extended in frame $t+d_t$ and $d_t$ is increased by $1$.

After the matching step, the iteration is repeated with the updated train set. The trajectories that either have been completed through matching or failed to be matched over the last $50$ iterations are removed from processing. In training step we set the learning rate to $0.00005$ and used the Adam optimizer \cite{Kingma2014-ln}, to ensure gradual learning of the partially altered train set. The solution was implemented in tensorflow and deployed on IBM power system with four P100 GPU's. The matching step was computed in parallel on 160 cores of the computational node with with train set modification, test set construction, and filtering of results performed by parallel process pools, Fig.~S7. 

For both recordings, network training and detection matching were first performed on half of the objects for $100$ iterations, after which the procedure was done for another $900$ iterations on all objects whose trajectories were yet unfinished.  Reducing the number of tracked objects allowed for faster training epochs in the initial $100$ iterations during which several trajectories were completed. The total of $1000$ iterations was completed during one week using one computation node for each recording.

\begin{figure*}[t]
  \centering
    \includegraphics[width=\textwidth]{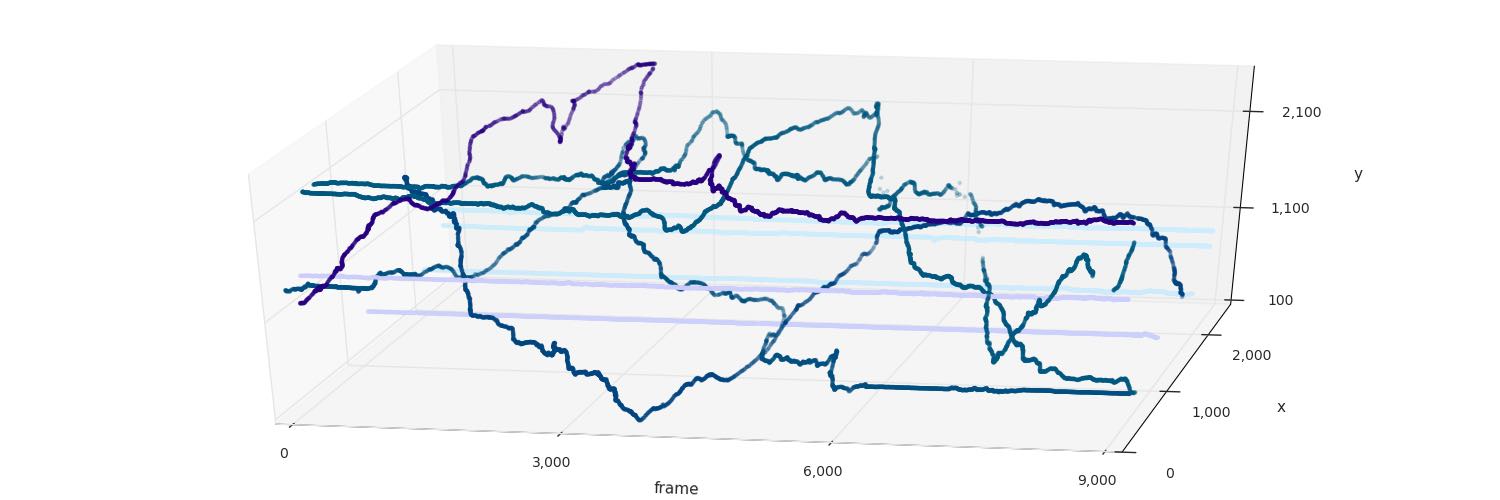}
    \caption{\figtitle{Tracking reveals a variety of movement dynamics within the hive.} Darker colors represent example trajectories of rapidly moving bees, while bright colors represent stationary bees.}\label{fig:4tracks}
\end{figure*}

\section*{Results}

To quantify the accuracy of our results in comparable format, we adopted the metrics used in the MOT16 benchmark \cite{MOT16}. Importantly however, in the context of a honeybee hive, ground truth detection of the trajectories would be prohibitively costly to obtain and thus we omit Multiple Object Tracking Accuracy (MOTA) and Multiple Object Tracking Precision (MOTP). Even so, we provide other track quality measures which capture the effectiveness of our solution. 

We used two tracking quality measures of MOT16 -- mostly tracked (MT) and mostly lost (ML) track -- that we quantified over the first 2 min and the entire 5 min of both recordings. According to MOT16, a target is mostly tracked if it is successfully tracked for at least $80\%$ of its life span. It is irrelevant for this measure whether the identity remains the same throughout the track. Mostly lost tracks are those only recovered for less than $20\%$ of the tested time span. Also, we did not consider as track discontinuities temporary identity changes after which the track comes back to its original identity. 

Trajectories counted as MT2 in Table~\ref{tab:results} are those tracked for at least $100\, {\rm s}$, MT5 are those tracked for at least 4 min, ML2 tracks are recovered for less than $10\, {\rm s} $, and ML5 for less than $30\, {\rm s} $. Within the trajectories not counted as MT5 we additionally assessed the sources of errors as detection error, identity swap, occlusion, or trajectory loss due to lack of detections that were recognized as the respective object.

We found $74\%$ and $71\%$ of trajectories mostly tracked over $2\, {\rm min}$ in recording 1 and 2, respectively. $46\%$ of trajectories were mostly tracked over the entire $5\, {\rm min} $ of both recordings. Few trajectories were mostly lost with $12\%$ and $16\%$ classified as ML5 in recording 1 and 2, respectively. Unsurprisingly, identity swaps among the highly similar honeybees were the predominant cause of tracking errors with $19\%$ and $30\%$ trajectories lost due to identity swap in both recordings, respectively. Trajectory loss is the second most frequent reason for tracking error. This type of error might result from the inaccuracy of the object recognition network, as well as from previous incorrect associations of tracks that might have incorrectly assigned the given object's track to another object in one of the previous iterations of the track matching algorithm leaving no possible detections to associate to this object's trajectory. Trajectory loss due to occlusions ($10\%$ and $3\%$ trajectories, respectively) might be caused by a dramatic change in object's visual cues, not captured in the train set as the object was occluded, as well as, if the occlusion lasts longer than $50$ frames the trajectory was stopped according to the algorithm design. Importantly, only a very small number of trajectories ($2-3\%$) were lost due to detection errors, corroborating the high accuracy of the object detection method.

The setting of this important biological system is substantially different than those considered in the MOT16 benchmark. The density of objects, number of object classes, visual variability of the scene, length and size of the recordings differentiate our study from the videos included in the benchmark. Nevertheless, our results according to the MT and ML measures ($22\%$ and $37\%$, respectively in MOT16) are both promising and substantially improved. Similarly, we are currently lacking ground truth detections to comprehensively estimate MOTA and MOTP measures. However, the low percentage of trajectory loss due to detection errors suggest that our detection accuracy is similar or better than the benchmark results.

We also note that our solution addresses several important challenges of tracking in dense environments such as rapid, irregular motion (MovieS1), body posture changes (MovieS2), temporary occlusions (MovieS3), or densely packed environments (MovieS4). Also, $70\%$ of objects tracked over the first $2\, {\rm min}$ and $46\%$ of objects tracked over $5\, {\rm min}$ represent a signficant  proportion of the densely packed comb (MovieS5 and MovieS6).  Supplementary movies are accessible until Dec 31, 2019 through the following link:\,\href {https://www.dropbox.com/sh/hxc529hyhd6mom4/AAAVv3Fjb8csbj9LdN4nU20za?dl=0}{{\bf https://tinyurl.com/y7lnesw4}} or by contacting the authors directly.

Currently limited to $5\, {\rm min}$, our solution can be straightforwardly parallelized across computational nodes and time segments, thus allowing for longer term tracking. Even within $5\, {\rm min}$, essential behavioral dynamics can be observed (Fig.~\ref{fig:4tracks}, S8, S9) enabling the quantitative analysis of collective properties of a naturally behaving honeybee colony at a single-organism resolution.  The capability to track without tags also facilitates the analysis of collective replicates and is relevant in systems where population fluctuations are important such as dividing cells. While we developed our tracking solution in the biologically-relevant and interesting system of a honeybee hive, we expect our method to easily generalize to other contexts.
\medskip
\subsubsection*{Acknowledgments}
Funding for this work was provided by the OIST Graduate University to ASM and GS. Additional funding was provided by KAKENHI grants 16H06209 and 16KK0175 from the Japan Society for the Promotion of Science to ASM. We are grateful to Yoann Portugal for assistance with colony maintenance and image acquisition.

\bibliography{ref}

\begin{thebibliography}{25}%
\makeatletter
\providecommand \@ifxundefined [1]{%
 \@ifx{#1\undefined}
}%
\providecommand \@ifnum [1]{%
 \ifnum #1\expandafter \@firstoftwo
 \else \expandafter \@secondoftwo
 \fi
}%
\providecommand \@ifx [1]{%
 \ifx #1\expandafter \@firstoftwo
 \else \expandafter \@secondoftwo
 \fi
}%
\providecommand \natexlab [1]{#1}%
\providecommand \enquote  [1]{``#1''}%
\providecommand \bibnamefont  [1]{#1}%
\providecommand \bibfnamefont [1]{#1}%
\providecommand \citenamefont [1]{#1}%
\providecommand \href@noop [0]{\@secondoftwo}%
\providecommand \href [0]{\begingroup \@sanitize@url \@href}%
\providecommand \@href[1]{\@@startlink{#1}\@@href}%
\providecommand \@@href[1]{\endgroup#1\@@endlink}%
\providecommand \@sanitize@url [0]{\catcode `\\12\catcode `\$12\catcode
  `\&12\catcode `\#12\catcode `\^12\catcode `\_12\catcode `\%12\relax}%
\providecommand \@@startlink[1]{}%
\providecommand \@@endlink[0]{}%
\providecommand \url  [0]{\begingroup\@sanitize@url \@url }%
\providecommand \@url [1]{\endgroup\@href {#1}{\urlprefix }}%
\providecommand \urlprefix  [0]{URL }%
\providecommand \Eprint [0]{\href }%
\providecommand \doibase [0]{http://dx.doi.org/}%
\providecommand \selectlanguage [0]{\@gobble}%
\providecommand \bibinfo  [0]{\@secondoftwo}%
\providecommand \bibfield  [0]{\@secondoftwo}%
\providecommand \translation [1]{[#1]}%
\providecommand \BibitemOpen [0]{}%
\providecommand \bibitemStop [0]{}%
\providecommand \bibitemNoStop [0]{.\EOS\space}%
\providecommand \EOS [0]{\spacefactor3000\relax}%
\providecommand \BibitemShut  [1]{\csname bibitem#1\endcsname}%
\let\auto@bib@innerbib\@empty
\bibitem [{\citenamefont {Crall}\ \emph {et~al.}(2018)\citenamefont {Crall},
  \citenamefont {Switzer}, \citenamefont {Oppenheimer}, \citenamefont
  {Ford~Versypt}, \citenamefont {Dey}, \citenamefont {Brown}, \citenamefont
  {Eyster}, \citenamefont {Gu{\'e}rin}, \citenamefont {Pierce}, \citenamefont
  {Combes},\ and\ \citenamefont {de~Bivort}}]{Crall2018-dk}%
  \BibitemOpen
  \bibfield  {author} {\bibinfo {author} {\bibfnamefont {J.~D.}\ \bibnamefont
  {Crall}}, \bibinfo {author} {\bibfnamefont {C.~M.}\ \bibnamefont {Switzer}},
  \bibinfo {author} {\bibfnamefont {R.~L.}\ \bibnamefont {Oppenheimer}},
  \bibinfo {author} {\bibfnamefont {A.~N.}\ \bibnamefont {Ford~Versypt}},
  \bibinfo {author} {\bibfnamefont {B.}~\bibnamefont {Dey}}, \bibinfo {author}
  {\bibfnamefont {A.}~\bibnamefont {Brown}}, \bibinfo {author} {\bibfnamefont
  {M.}~\bibnamefont {Eyster}}, \bibinfo {author} {\bibfnamefont
  {C.}~\bibnamefont {Gu{\'e}rin}}, \bibinfo {author} {\bibfnamefont {N.~E.}\
  \bibnamefont {Pierce}}, \bibinfo {author} {\bibfnamefont {S.~A.}\
  \bibnamefont {Combes}}, \ and\ \bibinfo {author} {\bibfnamefont {B.~L.}\
  \bibnamefont {de~Bivort}},\ }\href@noop {} {\bibfield  {journal} {\bibinfo
  {journal} {Science}\ }\textbf {\bibinfo {volume} {362}},\ \bibinfo {pages}
  {683} (\bibinfo {year} {2018})}\BibitemShut {NoStop}%
\bibitem [{\citenamefont {Mersch}\ \emph {et~al.}(2013)\citenamefont {Mersch},
  \citenamefont {Crespi},\ and\ \citenamefont {Keller}}]{Mersch2013-uf}%
  \BibitemOpen
  \bibfield  {author} {\bibinfo {author} {\bibfnamefont {D.~P.}\ \bibnamefont
  {Mersch}}, \bibinfo {author} {\bibfnamefont {A.}~\bibnamefont {Crespi}}, \
  and\ \bibinfo {author} {\bibfnamefont {L.}~\bibnamefont {Keller}},\
  }\href@noop {} {\bibfield  {journal} {\bibinfo  {journal} {Science}\ }\textbf
  {\bibinfo {volume} {340}},\ \bibinfo {pages} {1090} (\bibinfo {year}
  {2013})}\BibitemShut {NoStop}%
\bibitem [{\citenamefont {Seeley}(2009)}]{Seeley2009-yz}%
  \BibitemOpen
  \bibfield  {author} {\bibinfo {author} {\bibfnamefont {T.~D.}\ \bibnamefont
  {Seeley}},\ }\href@noop {} {\emph {\bibinfo {title} {The Wisdom of the Hive:
  the social physiology of honey bee colonies}}}\ (\bibinfo  {publisher}
  {Harvard University Press},\ \bibinfo {year} {2009})\BibitemShut {NoStop}%
\bibitem [{\citenamefont {Bozek}\ \emph {et~al.}(2017)\citenamefont {Bozek},
  \citenamefont {Hebert}, \citenamefont {Mikheyev},\ and\ \citenamefont
  {Stephens}}]{Bozek2017-sj}%
  \BibitemOpen
  \bibfield  {author} {\bibinfo {author} {\bibfnamefont {K.}~\bibnamefont
  {Bozek}}, \bibinfo {author} {\bibfnamefont {L.}~\bibnamefont {Hebert}},
  \bibinfo {author} {\bibfnamefont {A.~S.}\ \bibnamefont {Mikheyev}}, \ and\
  \bibinfo {author} {\bibfnamefont {G.~J.}\ \bibnamefont {Stephens}},\ }in\
  \href@noop {} {\emph {\bibinfo {booktitle} {Computer Vision and Pattern
  Recognition ({CVPR)}, 2017 {IEEE}}}}\ (\bibinfo {year} {2017})\BibitemShut
  {NoStop}%
\bibitem [{\citenamefont {Kratz}\ and\ \citenamefont
  {Nishino}(2010)}]{Kratz2010-kb}%
  \BibitemOpen
  \bibfield  {author} {\bibinfo {author} {\bibfnamefont {L.}~\bibnamefont
  {Kratz}}\ and\ \bibinfo {author} {\bibfnamefont {K.}~\bibnamefont
  {Nishino}},\ }in\ \href@noop {} {\emph {\bibinfo {booktitle} {Computer Vision
  and Pattern Recognition ({CVPR)}, 2010 {IEEE} Conference on}}}\ (\bibinfo
  {publisher} {ieeexplore.ieee.org},\ \bibinfo {year} {2010})\ pp.\ \bibinfo
  {pages} {693--700}\BibitemShut {NoStop}%
\bibitem [{\citenamefont {Ali}\ and\ \citenamefont
  {Dailey}(2009)}]{Ali2009-sf}%
  \BibitemOpen
  \bibfield  {author} {\bibinfo {author} {\bibfnamefont {I.}~\bibnamefont
  {Ali}}\ and\ \bibinfo {author} {\bibfnamefont {M.~N.}\ \bibnamefont
  {Dailey}},\ }in\ \href@noop {} {\emph {\bibinfo {booktitle} {Advanced
  Concepts for Intelligent Vision Systems}}},\ \bibinfo {series and number}
  {Lecture Notes in Computer Science},\ \bibinfo {editor} {edited by\ \bibinfo
  {editor} {\bibfnamefont {J.}~\bibnamefont {Blanc-Talon}}, \bibinfo {editor}
  {\bibfnamefont {W.}~\bibnamefont {Philips}}, \bibinfo {editor} {\bibfnamefont
  {D.}~\bibnamefont {Popescu}}, \ and\ \bibinfo {editor} {\bibfnamefont
  {P.}~\bibnamefont {Scheunders}}}\ (\bibinfo  {publisher} {Springer Berlin
  Heidelberg},\ \bibinfo {year} {2009})\ pp.\ \bibinfo {pages}
  {540--549}\BibitemShut {NoStop}%
\bibitem [{\citenamefont {Ge}\ \emph {et~al.}(2012)\citenamefont {Ge},
  \citenamefont {Collins},\ and\ \citenamefont {Ruback}}]{Ge2012-tz}%
  \BibitemOpen
  \bibfield  {author} {\bibinfo {author} {\bibfnamefont {W.}~\bibnamefont
  {Ge}}, \bibinfo {author} {\bibfnamefont {R.~T.}\ \bibnamefont {Collins}}, \
  and\ \bibinfo {author} {\bibfnamefont {R.~B.}\ \bibnamefont {Ruback}},\
  }\href@noop {} {\bibfield  {journal} {\bibinfo  {journal} {IEEE Trans.
  Pattern Anal. Mach. Intell.}\ }\textbf {\bibinfo {volume} {34}},\ \bibinfo
  {pages} {1003} (\bibinfo {year} {2012})}\BibitemShut {NoStop}%
\bibitem [{\citenamefont {Rodriguez}\ \emph {et~al.}(2011)\citenamefont
  {Rodriguez}, \citenamefont {Sivic}, \citenamefont {Laptev},\ and\
  \citenamefont {Audibert}}]{Rodriguez2011-qo}%
  \BibitemOpen
  \bibfield  {author} {\bibinfo {author} {\bibfnamefont {M.}~\bibnamefont
  {Rodriguez}}, \bibinfo {author} {\bibfnamefont {J.}~\bibnamefont {Sivic}},
  \bibinfo {author} {\bibfnamefont {I.}~\bibnamefont {Laptev}}, \ and\ \bibinfo
  {author} {\bibfnamefont {J.~Y.}\ \bibnamefont {Audibert}},\ }in\ \href@noop
  {} {\emph {\bibinfo {booktitle} {2011 International Conference on Computer
  Vision}}}\ (\bibinfo {year} {2011})\ pp.\ \bibinfo {pages}
  {1235--1242}\BibitemShut {NoStop}%
\bibitem [{\citenamefont {Henschel}\ \emph {et~al.}(2016)\citenamefont
  {Henschel}, \citenamefont {Leal-Taix{\'e}}, \citenamefont {Rosenhahn},\ and\
  \citenamefont {Schindler}}]{Henschel2016-zn}%
  \BibitemOpen
  \bibfield  {author} {\bibinfo {author} {\bibfnamefont {R.}~\bibnamefont
  {Henschel}}, \bibinfo {author} {\bibfnamefont {L.}~\bibnamefont
  {Leal-Taix{\'e}}}, \bibinfo {author} {\bibfnamefont {B.}~\bibnamefont
  {Rosenhahn}}, \ and\ \bibinfo {author} {\bibfnamefont {K.}~\bibnamefont
  {Schindler}},\ }\href@noop {} {\  (\bibinfo {year} {2016})},\ \Eprint
  {http://arxiv.org/abs/1607.07304} {arXiv:1607.07304 [cs.CV]} \BibitemShut
  {NoStop}%
\bibitem [{\citenamefont {{Bing Wang}}\ \emph {et~al.}(2017)\citenamefont
  {{Bing Wang}}, \citenamefont {{Gang Wang}}, \citenamefont {{Kap Luk Chan}},\
  and\ \citenamefont {{Li Wang}}}]{Bing_Wang2017-xf}%
  \BibitemOpen
  \bibfield  {author} {\bibinfo {author} {\bibnamefont {{Bing Wang}}}, \bibinfo
  {author} {\bibnamefont {{Gang Wang}}}, \bibinfo {author} {\bibnamefont {{Kap
  Luk Chan}}}, \ and\ \bibinfo {author} {\bibnamefont {{Li Wang}}},\
  }\href@noop {} {\bibfield  {journal} {\bibinfo  {journal} {IEEE Trans.
  Pattern Anal. Mach. Intell.}\ }\textbf {\bibinfo {volume} {39}},\ \bibinfo
  {pages} {589} (\bibinfo {year} {2017})}\BibitemShut {NoStop}%
\bibitem [{\citenamefont {Kim}\ \emph {et~al.}(2015)\citenamefont {Kim},
  \citenamefont {Li}, \citenamefont {Ciptadi},\ and\ \citenamefont
  {Rehg}}]{Kim2015-hr}%
  \BibitemOpen
  \bibfield  {author} {\bibinfo {author} {\bibfnamefont {C.}~\bibnamefont
  {Kim}}, \bibinfo {author} {\bibfnamefont {F.}~\bibnamefont {Li}}, \bibinfo
  {author} {\bibfnamefont {A.}~\bibnamefont {Ciptadi}}, \ and\ \bibinfo
  {author} {\bibfnamefont {J.~M.}\ \bibnamefont {Rehg}},\ }in\ \href@noop {}
  {\emph {\bibinfo {booktitle} {2015 {IEEE} International Conference on
  Computer Vision ({ICCV})}}}\ (\bibinfo {year} {2015})\ pp.\ \bibinfo {pages}
  {4696--4704}\BibitemShut {NoStop}%
\bibitem [{\citenamefont {Kuo}\ and\ \citenamefont
  {Nevatia}(2011)}]{Kuo2011-ie}%
  \BibitemOpen
  \bibfield  {author} {\bibinfo {author} {\bibfnamefont {C.}~\bibnamefont
  {Kuo}}\ and\ \bibinfo {author} {\bibfnamefont {R.}~\bibnamefont {Nevatia}},\
  }in\ \href@noop {} {\emph {\bibinfo {booktitle} {{CVPR} 2011}}}\ (\bibinfo
  {year} {2011})\ pp.\ \bibinfo {pages} {1217--1224}\BibitemShut {NoStop}%
\bibitem [{\citenamefont {Leal-Taix{\'e}}\ \emph {et~al.}(2016)\citenamefont
  {Leal-Taix{\'e}}, \citenamefont {Canton-Ferrer},\ and\ \citenamefont
  {{others}}}]{Leal-Taixe2016-wn}%
  \BibitemOpen
  \bibfield  {author} {\bibinfo {author} {\bibfnamefont {L.}~\bibnamefont
  {Leal-Taix{\'e}}}, \bibinfo {author} {\bibfnamefont {C.}~\bibnamefont
  {Canton-Ferrer}}, \ and\ \bibinfo {author} {\bibnamefont {{others}}},\
  }\href@noop {} {\bibfield  {journal} {\bibinfo  {journal} {Proc. IEEE}\ }
  (\bibinfo {year} {2016})}\BibitemShut {NoStop}%
\bibitem [{\citenamefont {Milan}\ \emph
  {et~al.}(2016{\natexlab{a}})\citenamefont {Milan}, \citenamefont
  {Rezatofighi}, \citenamefont {Dick}, \citenamefont {Reid},\ and\
  \citenamefont {Schindler}}]{Milan2016-gq}%
  \BibitemOpen
  \bibfield  {author} {\bibinfo {author} {\bibfnamefont {A.}~\bibnamefont
  {Milan}}, \bibinfo {author} {\bibfnamefont {S.~H.}\ \bibnamefont
  {Rezatofighi}}, \bibinfo {author} {\bibfnamefont {A.}~\bibnamefont {Dick}},
  \bibinfo {author} {\bibfnamefont {I.}~\bibnamefont {Reid}}, \ and\ \bibinfo
  {author} {\bibfnamefont {K.}~\bibnamefont {Schindler}},\ }\href@noop {} {\
  (\bibinfo {year} {2016}{\natexlab{a}})},\ \Eprint
  {http://arxiv.org/abs/1604.03635} {arXiv:1604.03635 [cs.CV]} \BibitemShut
  {NoStop}%
\bibitem [{\citenamefont {Ondruska}\ \emph {et~al.}(2016)\citenamefont
  {Ondruska}, \citenamefont {Dequaire}, \citenamefont {Wang},\ and\
  \citenamefont {Posner}}]{Ondruska2016-xy}%
  \BibitemOpen
  \bibfield  {author} {\bibinfo {author} {\bibfnamefont {P.}~\bibnamefont
  {Ondruska}}, \bibinfo {author} {\bibfnamefont {J.}~\bibnamefont {Dequaire}},
  \bibinfo {author} {\bibfnamefont {D.~Z.}\ \bibnamefont {Wang}}, \ and\
  \bibinfo {author} {\bibfnamefont {I.}~\bibnamefont {Posner}},\ }\href@noop {}
  {\  (\bibinfo {year} {2016})},\ \Eprint {http://arxiv.org/abs/1604.05091}
  {arXiv:1604.05091 [cs.LG]} \BibitemShut {NoStop}%
\bibitem [{\citenamefont {Ondruska}\ and\ \citenamefont
  {Posner}(2016)}]{Ondruska2016-dd}%
  \BibitemOpen
  \bibfield  {author} {\bibinfo {author} {\bibfnamefont {P.}~\bibnamefont
  {Ondruska}}\ and\ \bibinfo {author} {\bibfnamefont {I.}~\bibnamefont
  {Posner}},\ }\href@noop {} {\  (\bibinfo {year} {2016})},\ \Eprint
  {http://arxiv.org/abs/1602.00991} {arXiv:1602.00991 [cs.LG]} \BibitemShut
  {NoStop}%
\bibitem [{\citenamefont {Milan}\ \emph
  {et~al.}(2016{\natexlab{b}})\citenamefont {Milan}, \citenamefont
  {Leal-Taix\'{e}}, \citenamefont {Reid}, \citenamefont {Roth},\ and\
  \citenamefont {Schindler}}]{MOT16}%
  \BibitemOpen
  \bibfield  {author} {\bibinfo {author} {\bibfnamefont {A.}~\bibnamefont
  {Milan}}, \bibinfo {author} {\bibfnamefont {L.}~\bibnamefont
  {Leal-Taix\'{e}}}, \bibinfo {author} {\bibfnamefont {I.}~\bibnamefont
  {Reid}}, \bibinfo {author} {\bibfnamefont {S.}~\bibnamefont {Roth}}, \ and\
  \bibinfo {author} {\bibfnamefont {K.}~\bibnamefont {Schindler}},\ }\href
  {http://arxiv.org/abs/1603.00831} {\bibfield  {journal} {\bibinfo  {journal}
  {arXiv:1603.00831 [cs]}\ } (\bibinfo {year} {2016}{\natexlab{b}})},\ \bibinfo
  {note} {arXiv: 1603.00831}\BibitemShut {NoStop}%
\bibitem [{\citenamefont {Romero-Ferrero}\ \emph {et~al.}(2018)\citenamefont
  {Romero-Ferrero}, \citenamefont {Bergomi}, \citenamefont {Hinz},
  \citenamefont {Heras},\ and\ \citenamefont
  {de~Polavieja}}]{Romero-Ferrero2018-fb}%
  \BibitemOpen
  \bibfield  {author} {\bibinfo {author} {\bibfnamefont {F.}~\bibnamefont
  {Romero-Ferrero}}, \bibinfo {author} {\bibfnamefont {M.~G.}\ \bibnamefont
  {Bergomi}}, \bibinfo {author} {\bibfnamefont {R.}~\bibnamefont {Hinz}},
  \bibinfo {author} {\bibfnamefont {F.~J.~H.}\ \bibnamefont {Heras}}, \ and\
  \bibinfo {author} {\bibfnamefont {G.~G.}\ \bibnamefont {de~Polavieja}},\
  }\href@noop {} {\  (\bibinfo {year} {2018})},\ \Eprint
  {http://arxiv.org/abs/1803.04351} {arXiv:1803.04351 [cs.CV]} \BibitemShut
  {NoStop}%
\bibitem [{\citenamefont {Kabra}\ \emph {et~al.}(2013)\citenamefont {Kabra},
  \citenamefont {Robie}, \citenamefont {Rivera-Alba}, \citenamefont {Branson},\
  and\ \citenamefont {Branson}}]{Kabra2013-mg}%
  \BibitemOpen
  \bibfield  {author} {\bibinfo {author} {\bibfnamefont {M.}~\bibnamefont
  {Kabra}}, \bibinfo {author} {\bibfnamefont {A.~A.}\ \bibnamefont {Robie}},
  \bibinfo {author} {\bibfnamefont {M.}~\bibnamefont {Rivera-Alba}}, \bibinfo
  {author} {\bibfnamefont {S.}~\bibnamefont {Branson}}, \ and\ \bibinfo
  {author} {\bibfnamefont {K.}~\bibnamefont {Branson}},\ }\href@noop {}
  {\bibfield  {journal} {\bibinfo  {journal} {Nat. Methods}\ }\textbf {\bibinfo
  {volume} {10}},\ \bibinfo {pages} {64} (\bibinfo {year} {2013})}\BibitemShut
  {NoStop}%
\bibitem [{\citenamefont {Wario}\ \emph {et~al.}(2015)\citenamefont {Wario},
  \citenamefont {Wild}, \citenamefont {Couvillon}, \citenamefont {Rojas},\ and\
  \citenamefont {Landgraf}}]{Wario2015-fm}%
  \BibitemOpen
  \bibfield  {author} {\bibinfo {author} {\bibfnamefont {F.}~\bibnamefont
  {Wario}}, \bibinfo {author} {\bibfnamefont {B.}~\bibnamefont {Wild}},
  \bibinfo {author} {\bibfnamefont {M.~J.}\ \bibnamefont {Couvillon}}, \bibinfo
  {author} {\bibfnamefont {R.}~\bibnamefont {Rojas}}, \ and\ \bibinfo {author}
  {\bibfnamefont {T.}~\bibnamefont {Landgraf}},\ }\href@noop {} {\bibfield
  {journal} {\bibinfo  {journal} {Front. Ecol. Evol.}\ }\textbf {\bibinfo
  {volume} {3}} (\bibinfo {year} {2015})}\BibitemShut {NoStop}%
\bibitem [{\citenamefont {Gernat}\ \emph {et~al.}(2018)\citenamefont {Gernat},
  \citenamefont {Rao}, \citenamefont {Middendorf}, \citenamefont {Dankowicz},
  \citenamefont {Goldenfeld},\ and\ \citenamefont {Robinson}}]{Gernat2018-kw}%
  \BibitemOpen
  \bibfield  {author} {\bibinfo {author} {\bibfnamefont {T.}~\bibnamefont
  {Gernat}}, \bibinfo {author} {\bibfnamefont {V.~D.}\ \bibnamefont {Rao}},
  \bibinfo {author} {\bibfnamefont {M.}~\bibnamefont {Middendorf}}, \bibinfo
  {author} {\bibfnamefont {H.}~\bibnamefont {Dankowicz}}, \bibinfo {author}
  {\bibfnamefont {N.}~\bibnamefont {Goldenfeld}}, \ and\ \bibinfo {author}
  {\bibfnamefont {G.~E.}\ \bibnamefont {Robinson}},\ }\href@noop {} {\bibfield
  {journal} {\bibinfo  {journal} {Proc. Natl. Acad. Sci. U. S. A.}\ }\textbf
  {\bibinfo {volume} {115}},\ \bibinfo {pages} {1433} (\bibinfo {year}
  {2018})}\BibitemShut {NoStop}%
\bibitem [{\citenamefont {Szegedy}\ \emph {et~al.}(2015)\citenamefont
  {Szegedy}, \citenamefont {Vanhoucke}, \citenamefont {Ioffe}, \citenamefont
  {Shlens},\ and\ \citenamefont {Wojna}}]{Szegedy2015-xm}%
  \BibitemOpen
  \bibfield  {author} {\bibinfo {author} {\bibfnamefont {C.}~\bibnamefont
  {Szegedy}}, \bibinfo {author} {\bibfnamefont {V.}~\bibnamefont {Vanhoucke}},
  \bibinfo {author} {\bibfnamefont {S.}~\bibnamefont {Ioffe}}, \bibinfo
  {author} {\bibfnamefont {J.}~\bibnamefont {Shlens}}, \ and\ \bibinfo {author}
  {\bibfnamefont {Z.}~\bibnamefont {Wojna}},\ }\href@noop {} {\  (\bibinfo
  {year} {2015})},\ \Eprint {http://arxiv.org/abs/1512.00567} {arXiv:1512.00567
  [cs.CV]} \BibitemShut {NoStop}%
\bibitem [{\citenamefont {P{\'e}rez-Escudero}\ \emph
  {et~al.}(2014)\citenamefont {P{\'e}rez-Escudero}, \citenamefont
  {Vicente-Page}, \citenamefont {Hinz}, \citenamefont {Arganda},\ and\
  \citenamefont {de~Polavieja}}]{Perez-Escudero2014-se}%
  \BibitemOpen
  \bibfield  {author} {\bibinfo {author} {\bibfnamefont {A.}~\bibnamefont
  {P{\'e}rez-Escudero}}, \bibinfo {author} {\bibfnamefont {J.}~\bibnamefont
  {Vicente-Page}}, \bibinfo {author} {\bibfnamefont {R.~C.}\ \bibnamefont
  {Hinz}}, \bibinfo {author} {\bibfnamefont {S.}~\bibnamefont {Arganda}}, \
  and\ \bibinfo {author} {\bibfnamefont {G.~G.}\ \bibnamefont {de~Polavieja}},\
  }\href@noop {} {\bibfield  {journal} {\bibinfo  {journal} {Nat. Methods}\
  }\textbf {\bibinfo {volume} {11}},\ \bibinfo {pages} {743} (\bibinfo {year}
  {2014})}\BibitemShut {NoStop}%
\bibitem [{\citenamefont {Kingma}\ and\ \citenamefont
  {Ba}(2014)}]{Kingma2014-ln}%
  \BibitemOpen
  \bibfield  {author} {\bibinfo {author} {\bibfnamefont {D.}~\bibnamefont
  {Kingma}}\ and\ \bibinfo {author} {\bibfnamefont {J.}~\bibnamefont {Ba}},\
  }\href@noop {} {\  (\bibinfo {year} {2014})},\ \Eprint
  {http://arxiv.org/abs/1412.6980} {arXiv:1412.6980 [cs.LG]} \BibitemShut
  {NoStop}%
\bibitem [{\citenamefont {Ronneberger}\ \emph {et~al.}(2015)\citenamefont
  {Ronneberger}, \citenamefont {Fischer},\ and\ \citenamefont
  {Brox}}]{Ronneberger2015-bx}%
  \BibitemOpen
  \bibfield  {author} {\bibinfo {author} {\bibfnamefont {O.}~\bibnamefont
  {Ronneberger}}, \bibinfo {author} {\bibfnamefont {P.}~\bibnamefont
  {Fischer}}, \ and\ \bibinfo {author} {\bibfnamefont {T.}~\bibnamefont
  {Brox}},\ }in\ \href@noop {} {\emph {\bibinfo {booktitle} {Medical Image
  Computing and {Computer-Assisted} Intervention -- {MICCAI} 2015}}},\ \bibinfo
  {series and number} {Lecture Notes in Computer Science},\ \bibinfo {editor}
  {edited by\ \bibinfo {editor} {\bibfnamefont {N.}~\bibnamefont {Navab}},
  \bibinfo {editor} {\bibfnamefont {J.}~\bibnamefont {Hornegger}}, \bibinfo
  {editor} {\bibfnamefont {W.~M.}\ \bibnamefont {Wells}}, \ and\ \bibinfo
  {editor} {\bibfnamefont {A.~F.}\ \bibnamefont {Frangi}}}\ (\bibinfo
  {publisher} {Springer International Publishing},\ \bibinfo {year} {2015})\
  pp.\ \bibinfo {pages} {234--241}\BibitemShut {NoStop}%
\end{thebibliography}%
\clearpage
\onecolumngrid
\section*{Supplementary Material}

\setcounter{figure}{0} 
\renewcommand*{\thefigure}{S\arabic{figure}}

\begin{figure*}[h]
\begin{center}
  \includegraphics[height=4in]{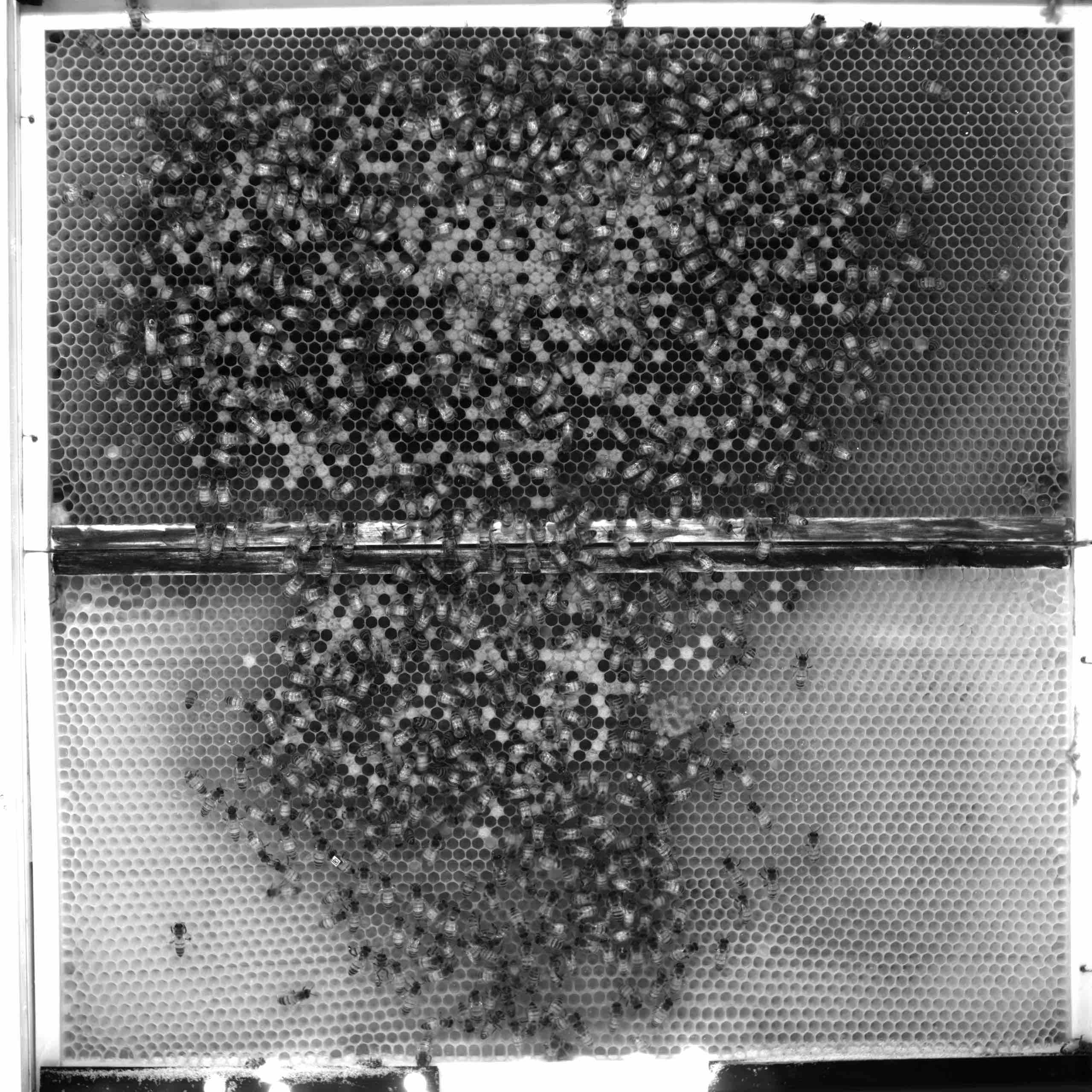}
  \\
  \vspace{0.1in}
  \includegraphics[height=3in]{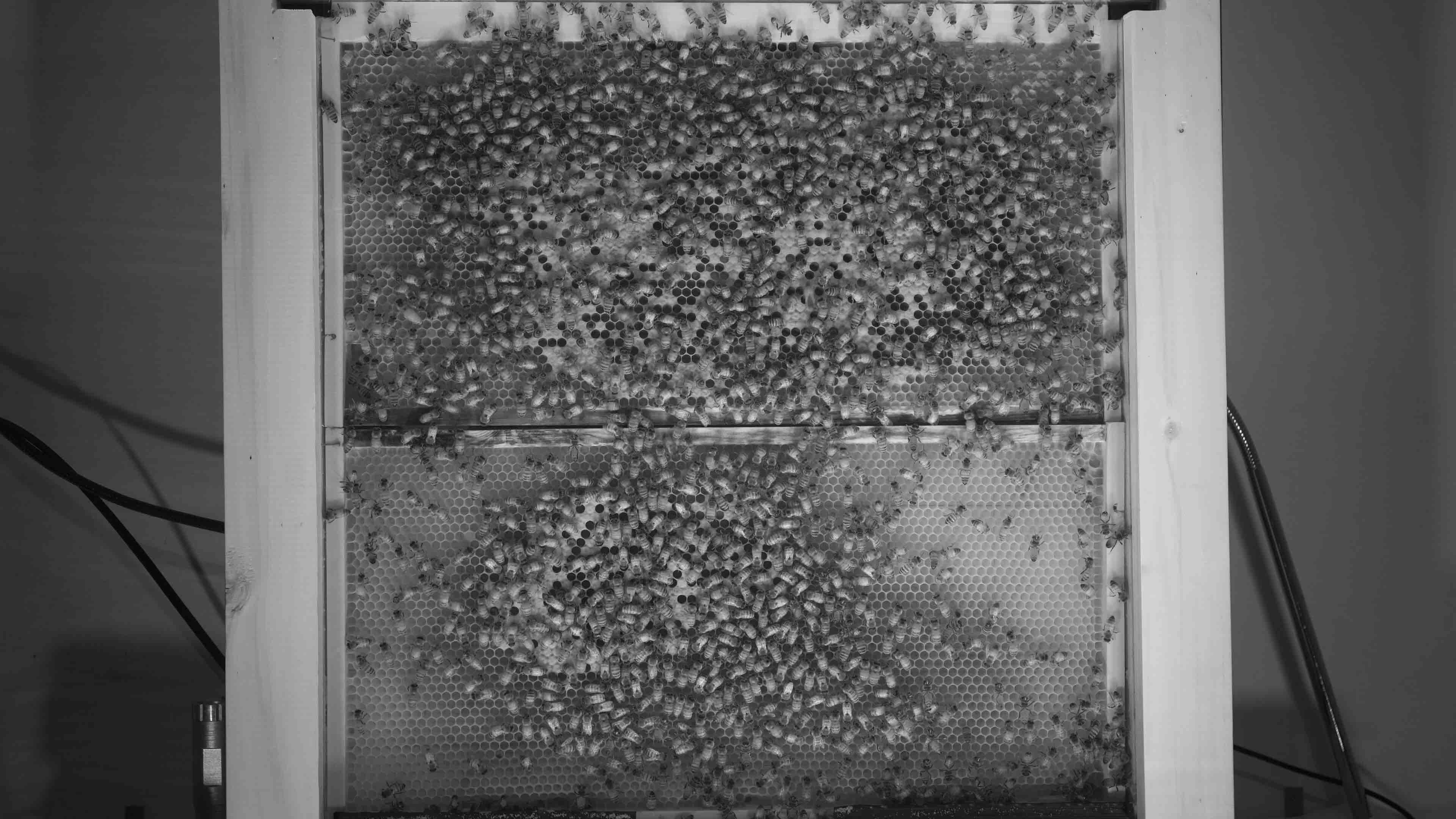}
\end{center}
   \caption{\figtitle{Example frames from recording 1 (upper image) and 2 (lower image).}}
\label{fig:S3frames}
\end{figure*}

\begin{figure*}[h]
\begin{center}
  \includegraphics[width=0.7\columnwidth]{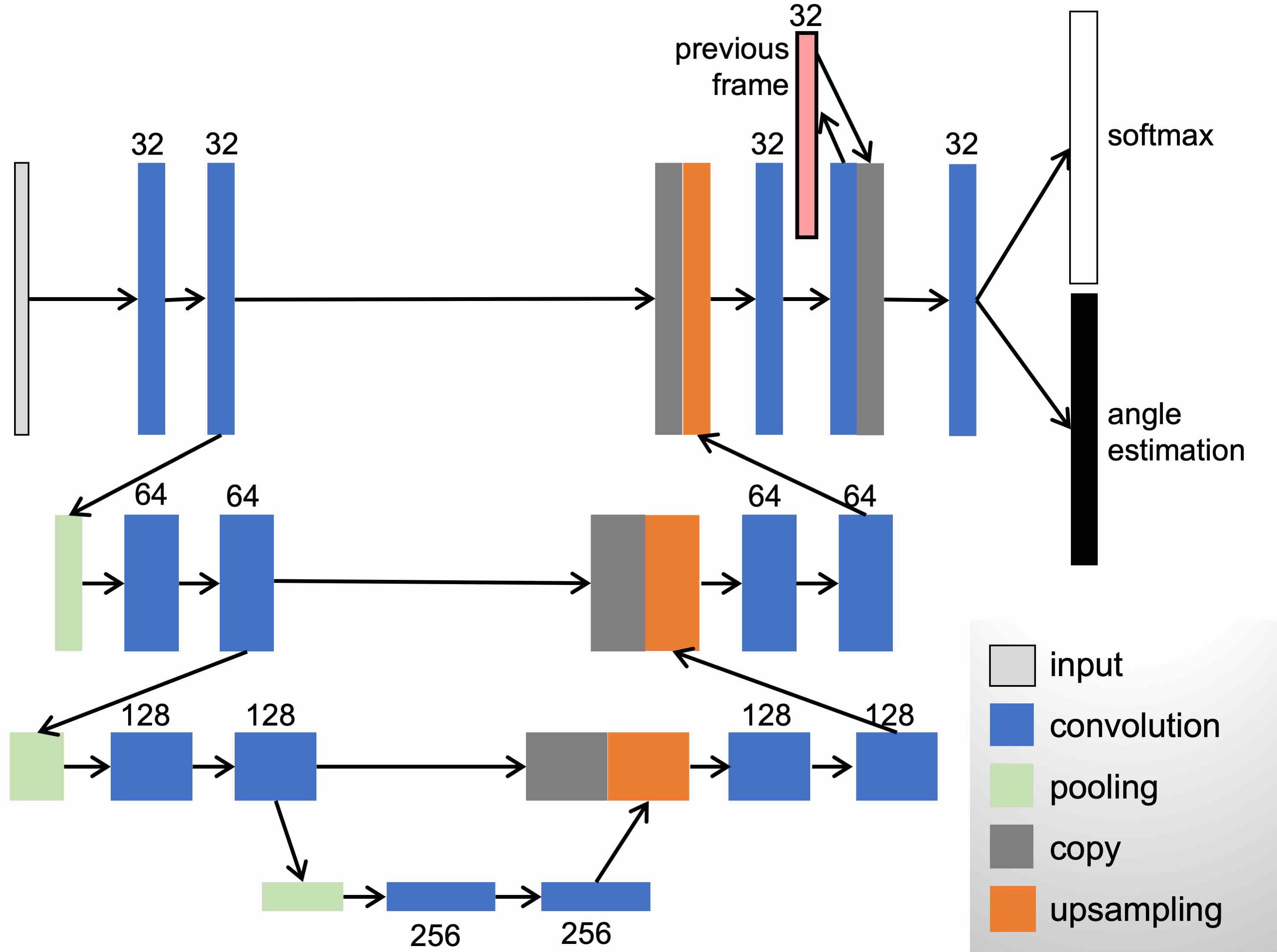}
\end{center}
   \caption{\figtitle{UNet architecture used for object detection.} The network was reduced to $\sim6\%$ of the parameters compared to the original UNet \cite{Ronneberger2015-bx}. A temporal component was added (pink box) to incorporate information from the previous video frame as a prior to improve the prediction of object positions in the following frame. An additional convolutional layer was added before the final two loss function layers -- softmax for class prediction and angle loss function as proposed in \cite{Bozek2017-sj}.}
\label{fig:S1unet}
\end{figure*}

\begin{figure*}
\begin{center}
  \includegraphics[height=2in]{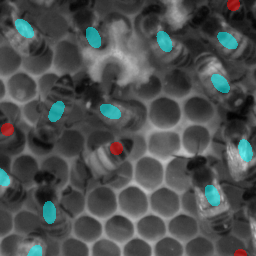}
  \includegraphics[height=2in]{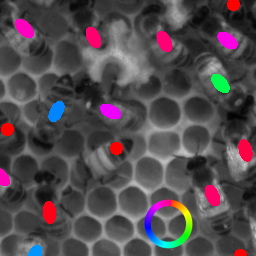}
  \end{center}
   \caption{\figtitle{Example labels and images used for object detection.} The Image size was $256\times256\, {\rm pixels}$, segmentation labels indicate object class (left panel) and orientation angle (right panel, angle color code indicated in the inserted circle). Segmentation labels are sized to $r1 = 6.7\, {\rm pixels}$ and $r2 =11.7\, {\rm pixels}$ for the semi-minor and semi-major axes of the ellipse-shaped label and $r = 6.7\, {\rm pixels}$ radius for the round-shaped label.}
\label{fig:S2segm}
\end{figure*}

\begin{figure*}
\begin{center}
  \includegraphics[width=0.8\columnwidth]{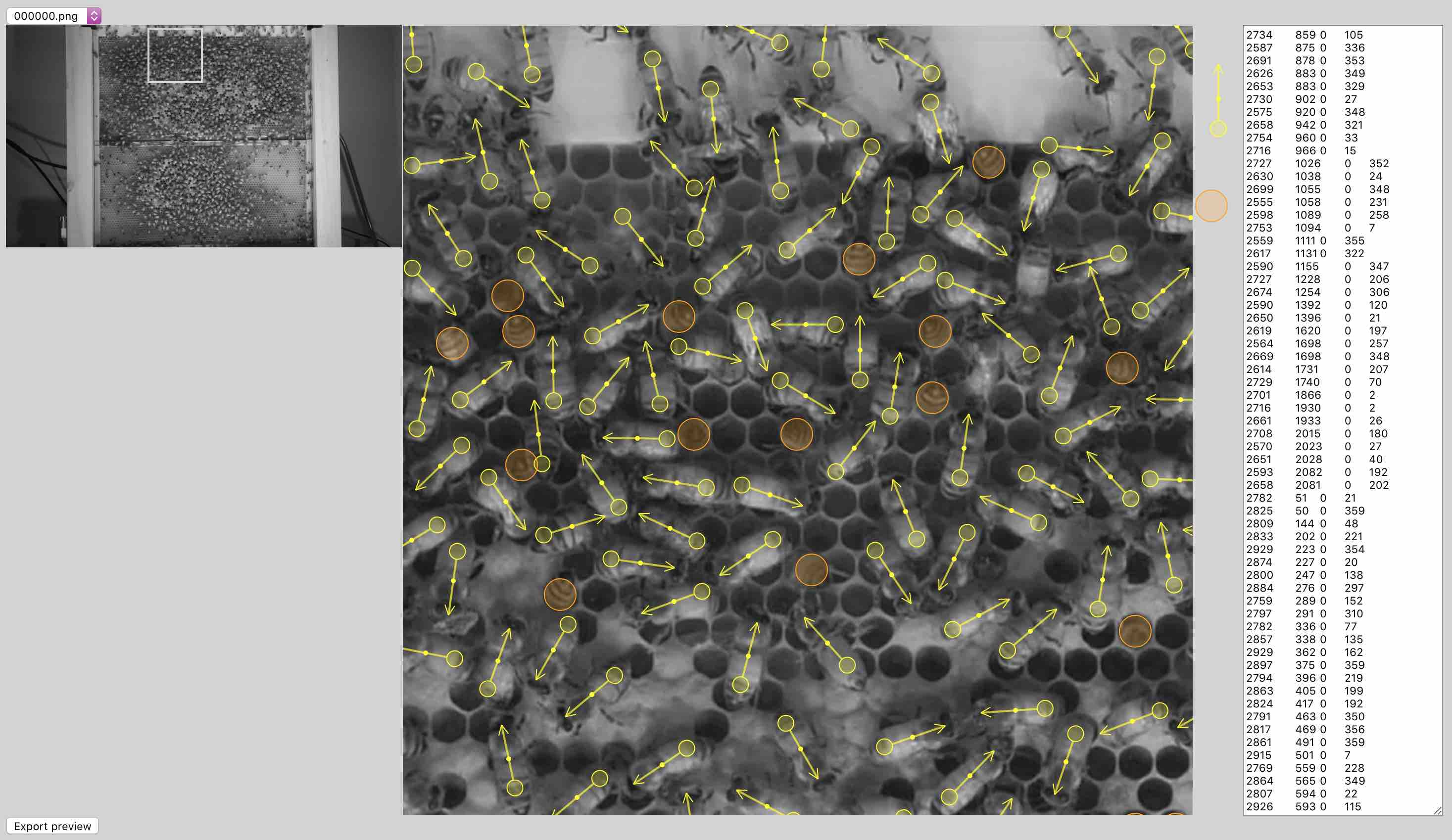}
\end{center}
   \caption{\figtitle{Labeling interface.} The interface was implemented in JavaScript. Labeling was done through dragging and dropping the yellow arrow and orange round symbols on the labelled image. The round end of the arrow allows for rotating it to align with the bee body axis. The entire video frame is labeled by scrolling with the use of square shaped zoom in the left image. An editable list of all annotations is displayed in the right panel.}
\label{fig:S4interface}
\end{figure*}

\begin{figure*}
\begin{center}
  \includegraphics[width=0.8\columnwidth]{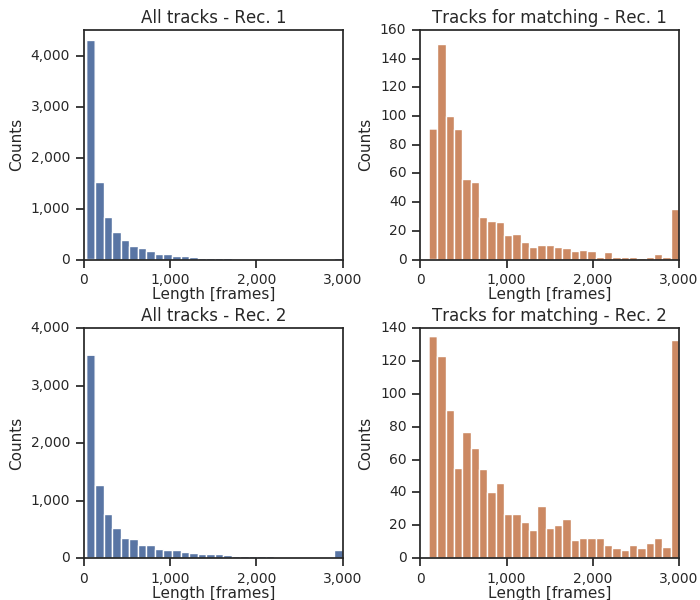}
\end{center}
   \caption{\figtitle{Distributions of track fragment lengths.} The Distributions of the lengths of track fragments obtained through position matching are shown for recording 1 (upper panels) and 2 (bottom panels). All tracks are included in the panels on the left, right-side panels show lengths of track fragments in the initial train set before the start of the track matching algorithm. Recording 2 contains a larger number of individuals with slower movement which resulted in a longer average track length compared to recording 1.}
\label{fig:S5tracklen}
\end{figure*}

\begin{figure*}
\begin{center}
  \includegraphics[height=1in]{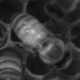}
  \includegraphics[height=1in]{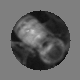}
  \includegraphics[height=1in]{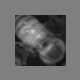}
  \\
  \vspace{0.02in}
  \includegraphics[height=1in]{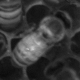}
  \includegraphics[height=1in]{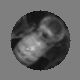}
  \includegraphics[height=1in]{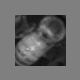}
  \\
  \vspace{0.02in}
  \includegraphics[height=1in]{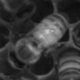}
  \includegraphics[height=1in]{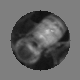}
  \includegraphics[height=1in]{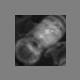}
  \\
  \vspace{0.02in}
  \includegraphics[height=1in]{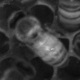}
  \includegraphics[height=1in]{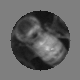}
  \includegraphics[height=1in]{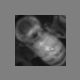}
\end{center}
   \caption{\figtitle{Augmentation operations.} All possible combinations of augmentation operations are shown on an example image (upper left image). These operations include flipping along 2 axes and round- or square-shaped masking of the outer parts of the image.}
\label{fig:S6augment}
\end{figure*}

\begin{figure*}
\begin{center}
  \includegraphics[width=0.8\columnwidth]{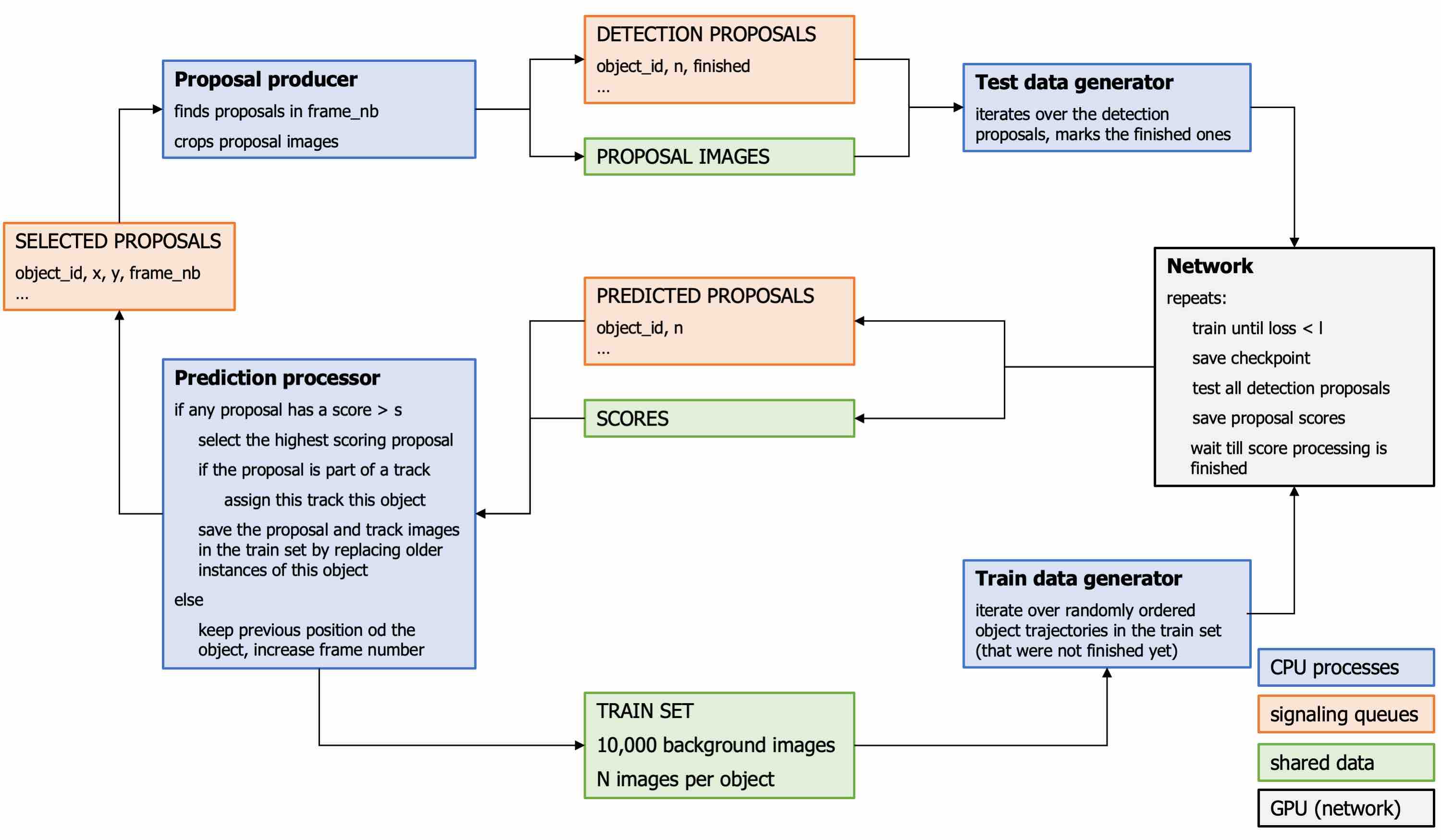}
\end{center}
   \caption{\figtitle{Training and tracking loop implementation.} Four process pools (blue boxes) are used for preparing images for the train and test and for processing of the network predictions. Queues (orange boxes) are used for process synchronization, images in the train and test sets are placed in shared data structures (green boxes).}
\label{fig:S7loop}
\end{figure*}

\begin{figure*}
\begin{center}
  \includegraphics[width=0.8\columnwidth]{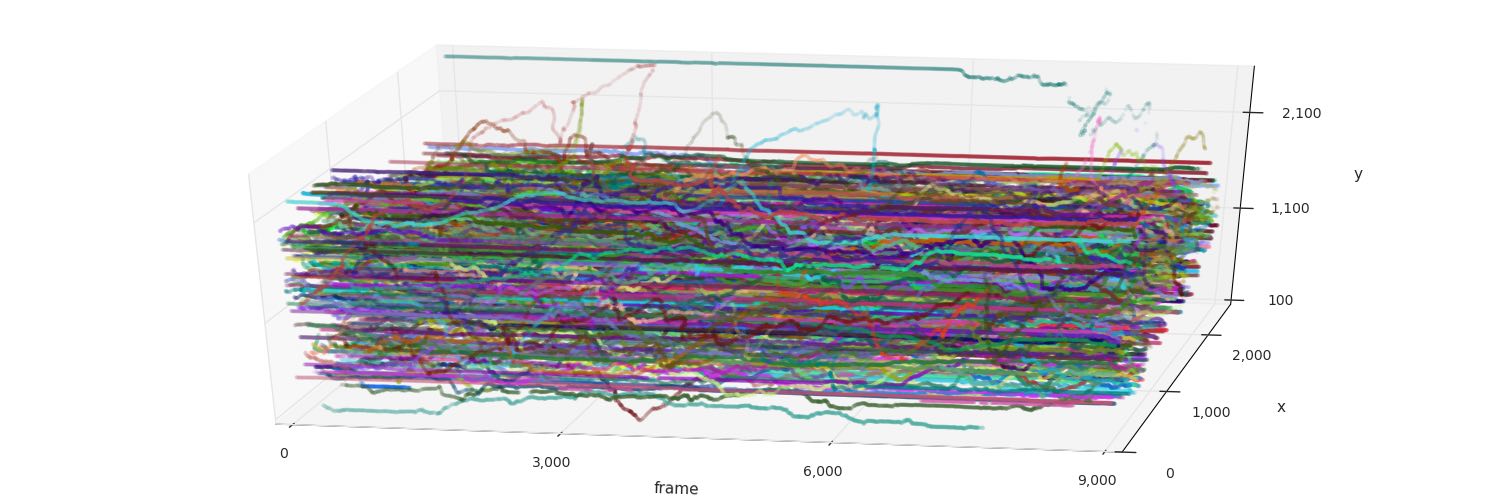} \\
  \vspace{0.02in}
  \includegraphics[width=0.8\columnwidth]{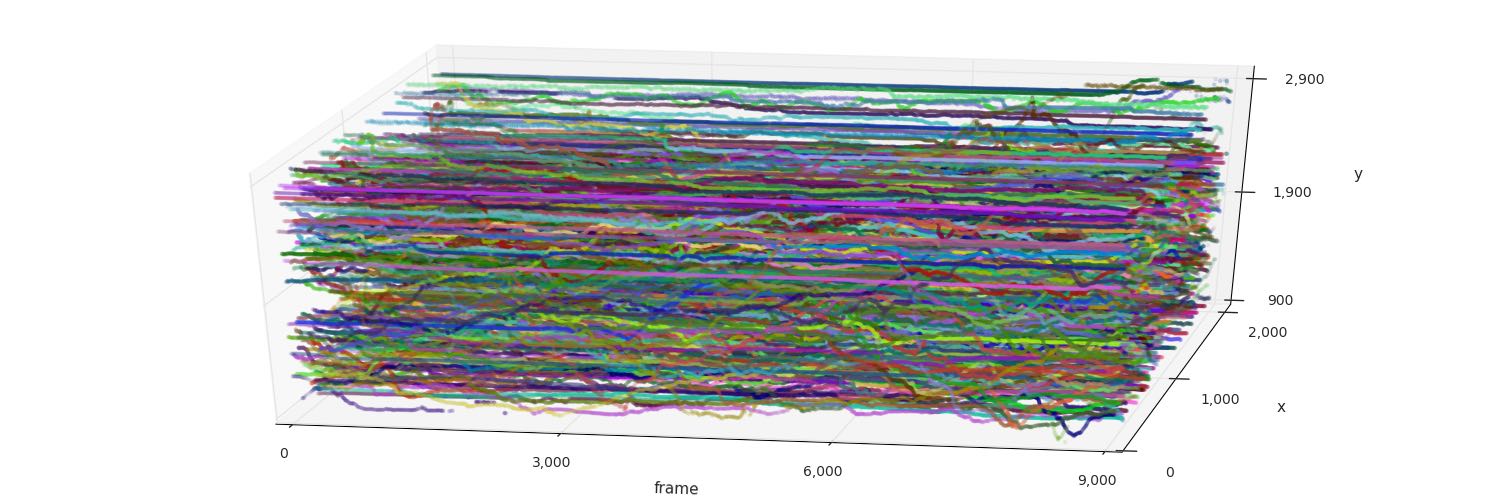}
\end{center}
   \caption{\figtitle{The reconstructed trajectories of 4 min and longer in recording 1 (upper plot) and 2 (bottom plot).}}
\label{fig:S8tracks}
\end{figure*}

\begin{figure*}
\begin{center}
  \includegraphics[width=0.8\columnwidth]{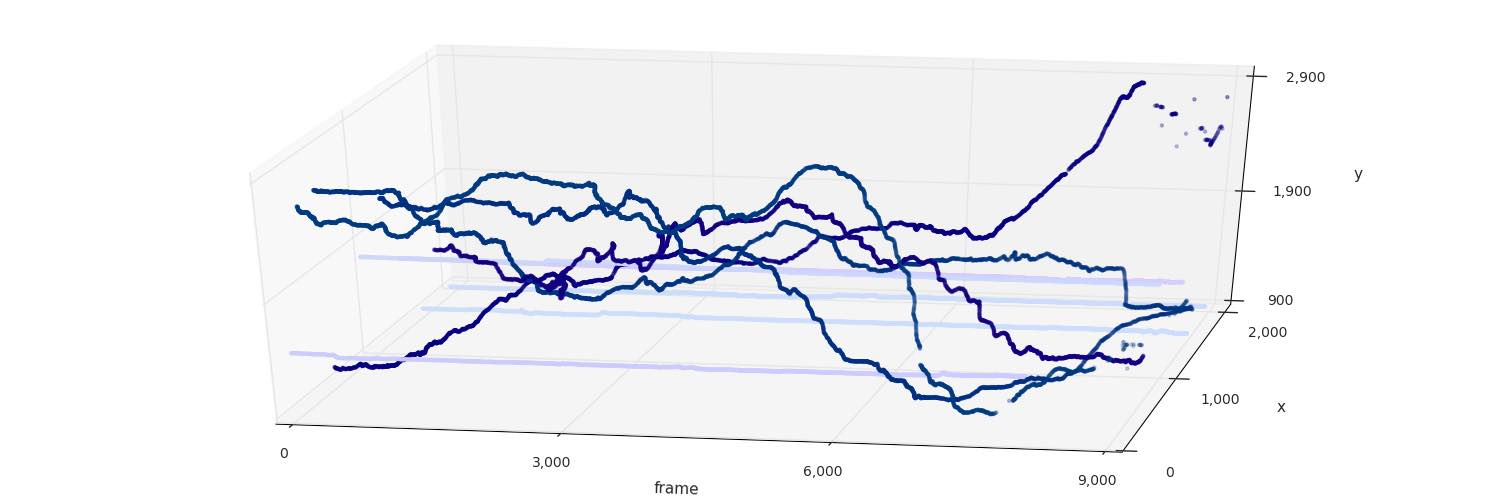} 
\end{center}
   \caption{\figtitle{Example trajectories of objects with different movement dynamics.} Analogous to Fig.~4, examples of fast moving and stationary object trajectories from recording 2 are shown in dark and bright colors, respectively.}
\label{fig:S9tracksP12}
\end{figure*}

\end{document}